\documentclass[letterpaper, 10 pt, conference]{ieeeconf}  

\IEEEoverridecommandlockouts                              

\usepackage{graphicx} 
\usepackage{amsmath} 

\usepackage{subcaption}
\usepackage{algorithmicx,algpseudocode}

\usepackage{url} 

\algnewcommand\algorithmicswitch{\textbf{switch}}
\algnewcommand\algorithmiccase{\textbf{case}}
\algnewcommand\algorithmicassert{\texttt{assert}}
\algnewcommand\Assert[1]{\State \algorithmicassert(#1)}%
\algdef{SE}[SWITCH]{Switch}{EndSwitch}[1]{\algorithmicswitch\ #1\ \algorithmicdo}{\algorithmicend\ \algorithmicswitch}%
\algdef{SE}[CASE]{Case}{EndCase}[1]{\algorithmiccase\ #1}{\algorithmicend\ \algorithmiccase}%
\algtext*{EndSwitch}%
\algtext*{EndCase}%

\title{\LARGE \bf
Towards CNN map representation and compression for camera relocalisation 
}

\author{Luis Contreras$^{1}$ and Walterio Mayol-Cuevas$^{2}$
\thanks{*This work was partially supported by CONACYT and the Secretaria de Educacion Publica, Mexico}%
\thanks{Department of Computer Science, University of Bristol, United Kingdom}%
\thanks{$^{1}${\tt\small cslact@my.bristol.ac.uk}}%
\thanks{$^{2}${\tt\small wmayol@cs.bris.ac.uk}}%
}

\begin{document}

\maketitle
\thispagestyle{empty}
\pagestyle{empty}

\begin{abstract}

This paper presents a study on the use of Convolutional Neural Networks for camera relocalisation and its application to map compression. We follow state of the art visual relocalisation results and evaluate the response to different data inputs. We use a CNN map representation and introduce the notion of map compression under this paradigm by using smaller CNN architectures without sacrificing relocalisation performance. We evaluate this approach in a series of publicly available datasets over a number of CNN architectures with different sizes, both in complexity and number of layers. This formulation allows us to improve relocalisation accuracy by increasing the number of training trajectories while maintaining a constant-size CNN.

\end{abstract}

\section{Introduction}
\label{sec:introduction}

Following our recent work on point cloud compression mapping via feature filtering in \cite{contreras:2015} and \cite{contreras:2017}, we aim to generate compact map representations useful for camera relocalisation through compact Convolutional Neural Networks (CNNs). This effort is motivated by the end-to-end approach of CNNs and in order to extend such to map compression. Overall, having a minimal map representation that enables later use is a meaningful question that underpins many applications for moving agents. In this work, we specifically explore a neural network architecture tested for the relocalisation task; we study the response of such architecture to different inputs -- e.g. color and depth images --, and the relocalisation performance of pre-trained neural networks in different tasks. 

Biologically inspired visual models have been proposed for a while \cite{harvey:1991}, \cite{milford:2004}. How humans improve learning after multiple training of the same view and how they filter useful information have also been an active field of study. One widely accepted theory of the human visual system suggests that a number of brain layers sequentially interact from the signal stimulus to the abstract concept \cite{dicarlo:2012}. Under this paradigm, the first layers -- connected directly to the input signal -- are a series of specialized filters that extract very specific features, while deeper layers infer more complex information by combining these features.

Finally, overfitting a neural network by excessive training with the same dataset is a well known issue; rather, here we study how the accuracy improves by revisiting the same area several times introducing new views to the dataset.

This paper is organized as follows. In Section \ref{sec:relatedwork} we discuss work related to convolutional neural networks and camera pose. Then, Section \ref{sect:cnnmap} introduces the notion of CNN map representation and compression. The CNN architectures used in the relocalisation task are then introduced in Section \ref{sect:cnn}, where we describe their architecture. Experimental results are presented in Section \ref{sect:experiments}. Finally, we  outline our discussion and conclusions.

\section{Related Work}
\label{sec:relatedwork}

Even though neural networks are not a novel concept, due to the increase in computational power, their popularity has grown in recent years \cite{bengio:courville:2012} \cite{bengio:courville:2013}. Related to map compression, dimensionality reduction through neural networks was first discussed in \cite{hinton:2006}. In \cite{chatfield:simonyan:2011} an evaluation to up-to-date data encoding algorithms for object recognition was presented, and it was extended in \cite{chatfield:simonyan:2014} to introduce the use of Convolutional Neural Networks for the same task.

\cite{agrawal:2015} introduced the idea of egomotion in CNN training by concatenating the output of two parallel neural networks with two different views of the same image; at the end, this architecture learns valuable features independent of the point of view.

In \cite{jarrett:2009}, the authors concluded that sophisticated architectures compensate for lack of training. \cite{garg:2016} explore this idea for single view depth estimation where they present a stereopsis based auto-encoder that uses few instances on the KITTI dataset. Then, \cite{eigen:2014}, \cite{li:shen:2015}, and \cite{liu:shen:2015} continued studying the use of elaborated CNN architectures for depth estimation.

Moving from depth to pose estimation was the next logical step. One of the first 6D camera pose regressors was presented in \cite{Gee:Mayol:2012} via a general regression NN (GRNN) with synthetic poses. More recently, PoseNet is presented in \cite{kendall:grimes:2015}, where they regress the camera pose using a CNN model. In the same sense, \cite{clark:2017} presented VidLoc, where they improve PoseNet results in offline video sequences by adding a biderctional RNN that takes advantage of the temporal information in the camera pose problem. This idea is also explored in \cite{long:kneip:2016} for image matching via training a CNN for frame interpolation through video sequences. 

\section{Map Representation as a Regression Function} \label{sect:cnnmap}

From a human observer point of view, it is common to think of spatial relationships among elements in space to build maps; for this reason, metric, symbolic, and topological are widely used map representations (such as probabilistic \cite{cummins:2008}, topological \cite{angeli:2008}, and metric and topological \cite{lim:frahm:2012} map representations). However, other less intuitive map representation have been proposed -- e.g. \cite{milford:2012} defines a map as a collection of images and uses image batch matching to find the current position in the map. 

Overall, it can be argued that the map representation needs not conform to a \textit{single} representation type, and that the task and other constraints can lead to different manners in which a map can be represented. Ultimately, for a robotic agent, maps are likely built to be explored or, more generally, re-explored. Thus, it is highlighted once more that relocalisation is a good measure of map effectiveness. In this context, the actual map representation used is less relevant as long as it allows relocalisation in a specific scene or task; therefore, we propose a mapping process based on Convolutional Neural Network, or CNN, to address the camera relocalisation problem.

\begin{figure}[h]
	\centering
	\includegraphics[width=0.45\textwidth]{{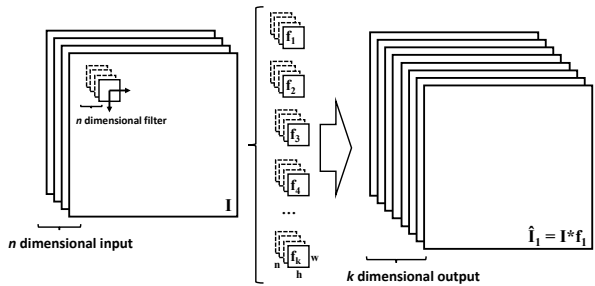}}
	\caption{Convolutional Neural Network (CNN) elements. It consist of an input $I$, a series of filters $f_k$, and its associated output $\hat{I}_k$ as a result of the convolution $I*f_k$. The filters' depth $n$ depends on the number of input channels.}
	\label{fig:cnnstruct}
\end{figure}

A CNN can be considered as a filter bank where the filters' weights are such that they minimize the error between an expected output and the system response to a given input. Figure \ref{fig:cnnstruct} shows the elements from one layer to the next in a typical CNN architecture -- a more detailed CNN implementation can be found in specialized works such as \cite{srinivas:2016} and \cite{nielsen:2015}. From the figure, for a given input $I$ and a series of $k$ filters $f_k$, it is generated an output $\hat{I}_k = I*f_k$, where $*$ represents the convolution operator (hence, this layer is also called \textit{convolutional} layer), where the filters $f_k$ can be initialized randomly or with pre-trained weights in a different task. It is important to notice the direct relationship between the input channels and the filters' depth among consecutive layers; it makes possible to work with different n-dimensional inputs just by adjusting the first convolutional layer depth.

As a result, in this work we represent a map as a regression function $\hat{p} = cnn(I)$, where each element of the population is formed by an input $I$ and its associated output $p$ (e.g. an RGB image and the 6-DoF camera pose, respectively). The parameters in the regressor $cnn$ are optimised from a population sample; the more representative the sample, the more accurate the model \cite{james:witten:hastie:tibshirani:2014}. 

\subsection{CON-POCO: CNN Map Compression} 

The notion of compression using a regression function as a map representation is introduced as follows. First, a population sample is defined as a collection of elements $(I,p)$ that represents a sensor's travelled trajectory and, further, this collection can be divided in training and testing sets. From the training set, a regressor $\hat{p} = cnn(I)$ is proposed such that it minimises the error $|p - \hat{p}|$ over the test set. 

This regressor, once defined, will have constant size, and should improve its performance while increasing the training set size (e.g. by generating more training trajectories) without increasing the regressor size itself. The compact map representation under the CNN paradigm is then stated as the problem of finding an optimal model cnn(I) that keep minimum relocalisation error values given a population sample.

\section{The Relocalisation CNN} \label{sect:cnn}

To evaluate the CNN map representation in the relocalisation task, we test several CNN architectures of the form $\hat{p} = cnn(I)$, where $I$ is an input image and the expected output is a 6-DoF pose $p = [x, q]$, with $x$ as the spatial position and $q$ as the orientation in quaternion form. We use PoseNet loss function, as described in \cite{kendall:grimes:2015}, that has the form:

$$loss(I)=\Vert\hat{x}-x\Vert_2+\beta\left\Vert\hat{q}-\frac{q}{\Vert q \Vert}\right\Vert_2$$

where PoseNet is based on the GoogLeNet arquitecture \cite{szegedy:2014}, and $\beta$ is a scale factor. As a reference, GoogLeNet has 7 million parameters but with a more elaborated architecture \cite{emer:2017}; in contrast, in this work to evaluate the relocalisation performance with respect to the CNN size, we only vary the number of convolutional layers and no interaction among them is introduced. Learning the external parameters in a CNN is time consuming because there is not really an optimal approach to this task; for this reason, the use of generic representations has been proposed such as in \cite{razavian:2014}, where the models trained in one task can be used in another, a process known as \textit{transfer learning}. Thus, we tested several architectures using a series of pre-trained models on the ImageNet dataset \cite{deng:dong:2009} and implemented in the MatConvNet platform \cite{vedaldi:2015}, as detailed bellow.

First, we use a relatively small CNN architectures with different complexities, as in \cite{chatfield:simonyan:2014}, with eight layers: five convolutional and three fully-connected. We use three implementations: a fast architecture (VGG-F) with 61 million parameters (considering an RGB input), where the first convolutional layer has a four pixel stride; a medium architecture (VGG-M) and 100 million parameters, where a smaller stride and a smaller filter size with a bigger depth are used in the first convolutional layer, and bigger filters' depths are used in the remaining convolutional layers. Finally, we study a slow architecture (VGG-S), with a similar architecture and number of parameters as in the VGG-M, but with a smaller stride in the second convolutional layer. 

Moreover, we evaluated two long CNN architectures, as in \cite{simonyan:2014}, one with 16 layers (13 convolutional and three fully connected layers with 138 million parameters) or VGG-16, and the other with 19 layers (16 convolutional and three fully connected layers with a total of 144 million parameters), referred as VGG-19. We introduce a couple of changes to these networks as follows: the dimension in the first layer depends on the input \textit{n}; in addition, the final fully-connected layer size changes to the pose vector length (i.e. from 4096 to 7). 

\begin{figure}[h]
\centering
	\begin{subfigure}{.1125\textwidth}
		\centering
		\includegraphics[bb = 0 0 640 360, width=1\textwidth]{{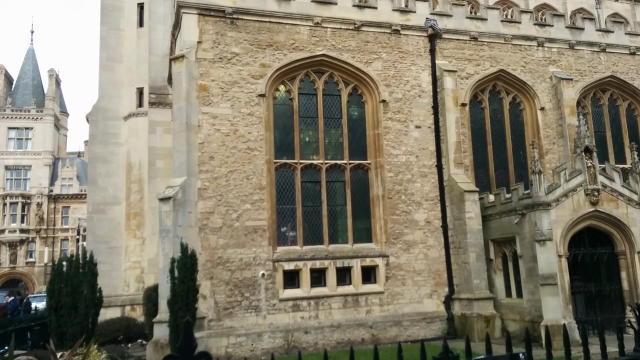}}
		\caption{}
		\label{fig:posenet1}
	\end{subfigure}
	\begin{subfigure}{.1125\textwidth}
  		\centering
  		\includegraphics[bb = 0 0 640 360, width=1\textwidth]{{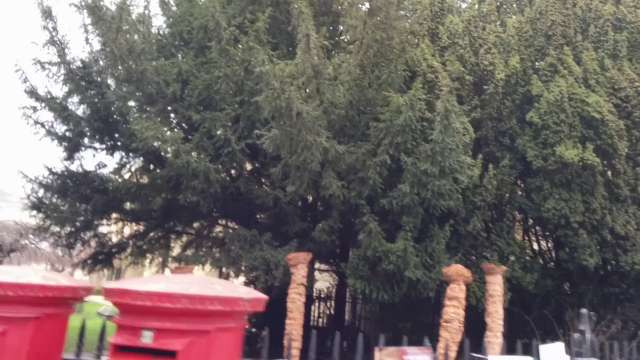}}
  		\caption{}
  		\label{fig:posenet2}
	\end{subfigure}
	\begin{subfigure}{.1125\textwidth}
  		\centering
  		\includegraphics[bb = 0 0 640 360, width=1\textwidth]{{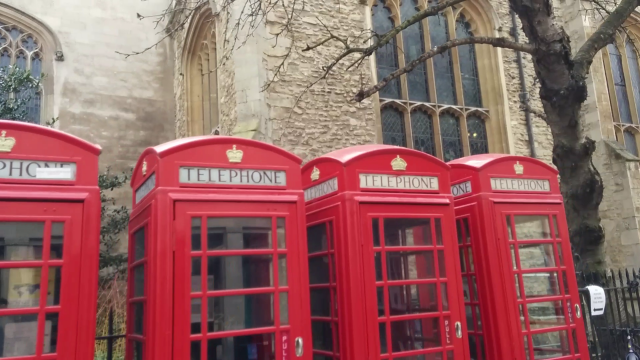}}
  		\caption{}
  		\label{fig:posenet3}
	\end{subfigure}
	\begin{subfigure}{.1125\textwidth}
  		\centering
  		\includegraphics[bb = 0 0 640 360, width=1\textwidth]{{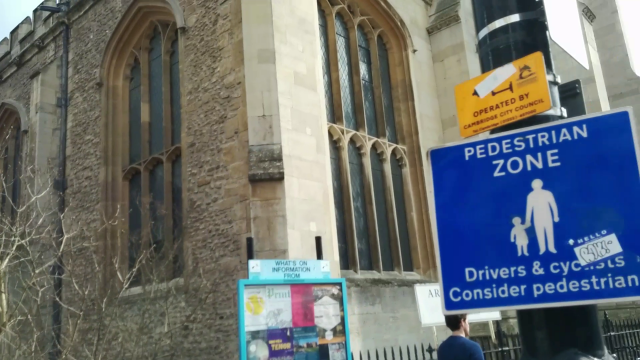}}
  		\caption{}
  		\label{fig:posenet4}
	\end{subfigure}
	\caption{Typical views from the St Marys Church sequence dataset \cite{kendall:grimes:2015b}.}
	\label{fig:posenet}
\end{figure}

To demonstrate the impact of smaller architectures in the relocalisation problem, we evaluate their performance in the St Marys Church sequence, a large scale outdoor scene \cite{kendall:grimes:2015b}, with 1487 training and 530 testing frames, as shown in Figure \ref{fig:posenet} and we only use RGB information as input, and pre-processing the original input by cropping the central area and resizing it, generating arrays of size 224x224. We compare their performance against PoseNet, as reported in \cite{kendall:grimes:2015}.

The training process is described next. First, we use the fast implementation (VGG-F) to find a valid range for the hyper-parameters involved in the CNN response, namely the batch size, the learning rate, and the weight decay. Then, with this valid hyper-parameters range, we perform a similar experiment using the proposed CNN architectures for the relocalisation task in the same dataset (the St Marys Church sequence) to find a valid hyper-parameters range in all architectures. We evaluate twenty hyper-parameters combination per architecture for 250 epochs.

Given the best hyper-parameters range from the process described above (batch size of 30, weight decays of 5E-01, and learning rates from 1E-06 to 1E-10), we perform an evaluation of the relocalisation performance using smaller CNN architectures than that in the original PoseNet. In general, larger architectures present better performance; however, none of them outperforms the baseline, as described bellow.

\begin{figure}[ht]
	\centering
	\includegraphics[width=0.45\textwidth]{{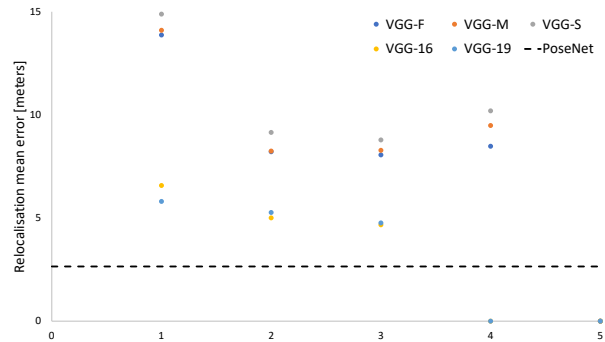}}
	\caption{Relocalisation performance to different CNN architectures in the St Marys Church dataset \cite{kendall:grimes:2015b}, where five parameter combinations were used per architecture during 500 training epochs: batch size of 30, weight decays of 5E-01, and learning rates from 1E-06 to 1E-10, where the position error is defined as the Euclidean distance between camera positions $\hat{x}$, $e_p = \Vert x - \hat{x}\Vert$. In dotted lines are the results for PoseNet as in \cite{kendall:grimes:2015}.}
	\label{fig:cnnposenetpos}
\end{figure}

Figure \ref{fig:cnnposenetpos} shows the error in position, defined as the Euclidean distance between the expected position $x$ and the CNN response $\hat{x}$, $e_p = \Vert x - \hat{x}\Vert$; here, the minimum error obtained was 4.76 meters using the VGG-19 CNN architecture (for comparison, PoseNet's error is 2.65 meters). Then, the smallest error in orientation -- in this work, it is given by the angle between the two quaternion vectors $ e_a = cos^{-1}(<q, \hat{q}>)$ -- is 10.40$^{\circ}$, also given by the VGG-19 CNN architecture, and again PoseNet (whose error is 4.24$^{\circ}$) outperforms in this scenario.

\begin{figure}[ht]
	\centering
	\includegraphics[width=0.45\textwidth]{{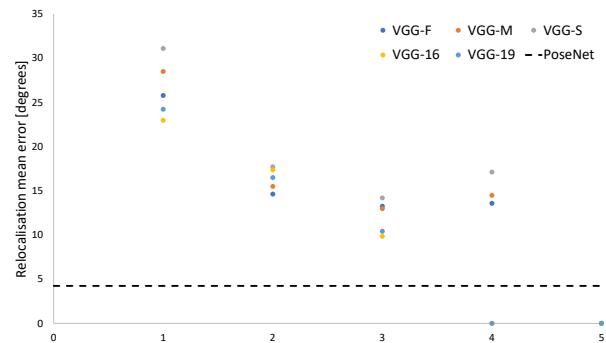}}
	\caption{Orientation error, as the angle between quaternion orientations $ e_a = cos^{-1}(<q, \hat{q}>)$. Parameters are as in to Figure \ref{fig:cnnposenetpos}.}
	\label{fig:cnnposenetor}
\end{figure}

\begin{figure}[ht]
\centering
	\begin{subfigure}{.1125\textwidth}
		\centering
		\includegraphics[bb = 0 0 640 480, width=1\textwidth]{{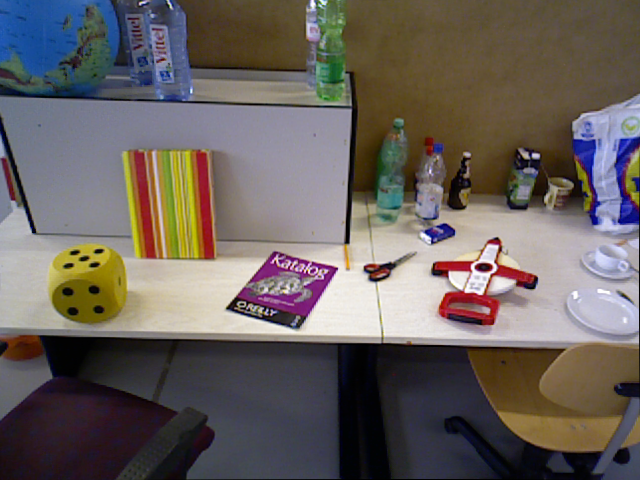}}
		\caption{}
		\label{fig:tum1}
	\end{subfigure}
	\begin{subfigure}{.1125\textwidth}
  		\centering
  		\includegraphics[bb = 0 0 640 480, width=1\textwidth]{{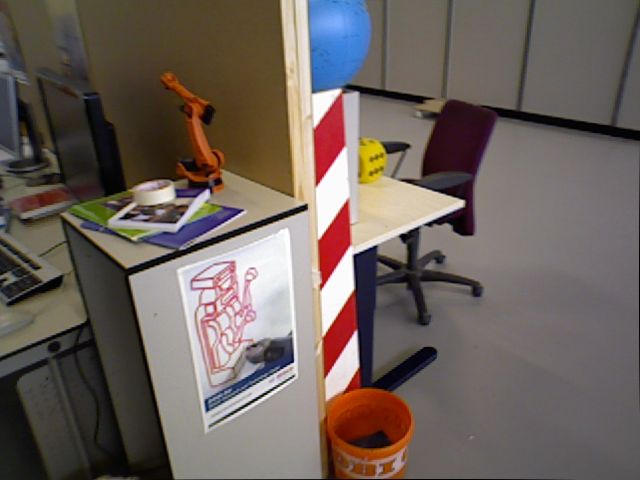}}
  		\caption{}
  		\label{fig:tum2}
	\end{subfigure}
	\begin{subfigure}{.1125\textwidth}
  		\centering
  		\includegraphics[bb = 0 0 640 480, width=1\textwidth]{{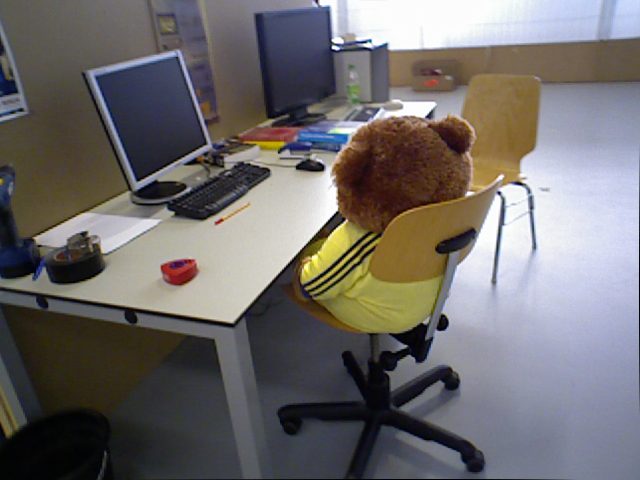}}
  		\caption{}
  		\label{fig:tum3}
	\end{subfigure}
	\begin{subfigure}{.1125\textwidth}
  		\centering
  		\includegraphics[bb = 0 0 640 480, width=1\textwidth]{{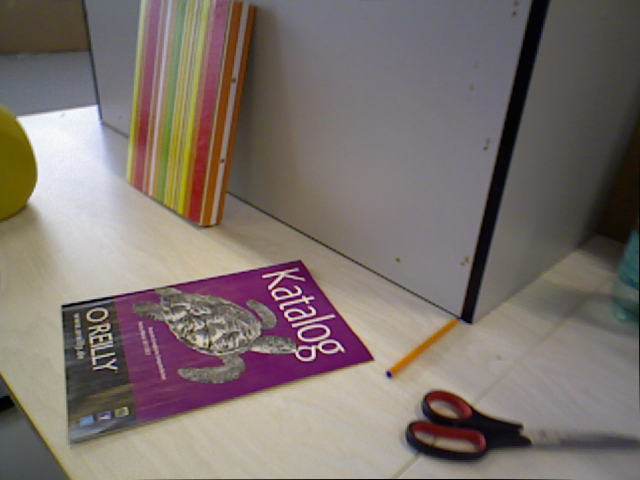}}
  		\caption{}
  		\label{fig:tum4}
	\end{subfigure}
	\caption{Some scenes in the TUM long office sequence \cite{sturm:2012}.}
	\label{fig:tum}
\end{figure}

Given that a regression model performance increases with the number of training samples, we perform a similar evaluation in a larger dataset to assess their effect in the CNN map representation. In particular, we tested these implementations in the TUM's long household and office sequence \cite{sturm:2012} -- a texture and structure rich scene, with 21.5m in 87.09s (2585 frames), and a validation dataset with 2676 extra frames, as can be seen in Figure \ref{fig:tum}.

\begin{figure}[ht]
	\centering
	\includegraphics[width=0.45\textwidth]{{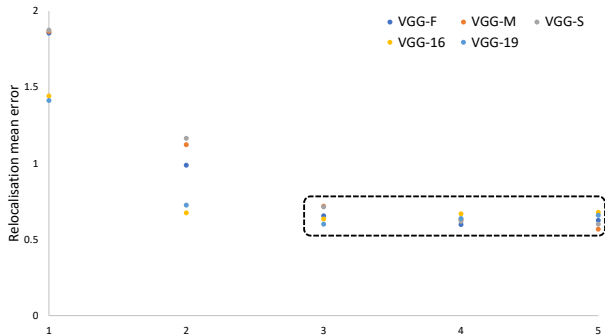}}
	\caption{Relocalisation response, as the distance between pose vectors $e = \Vert p - \hat{p} \Vert$, among different CNN architectures in the TUM long office dataset \cite{sturm:2012}; each point represents a hyper-parameter combination with a batch size of 30, weight decays of 5E-01, and learning rates from 1E-06 to 1E-10. For the highlighted combinations, there is no significant difference in the performance among short and long architectures in this dataset.}
	\label{fig:cnntumerr}
\end{figure}

Figure \ref{fig:cnntumerr} shows the relocalisation performance for different CNN architectures -- to discover the difference in performance between architectures, we use the Euclidean distance between pose vectors $e = \Vert p - \hat{p} \Vert$ as error metric, where $p = [x, q]$, and $\hat{p}$ is the estimated camera pose. It is observed that, for some parameter combinations, there is no significant difference in performance between short and long architectures, and therefore short architectures (i.e. compact map representations) may be used in the relocalisation task in such cases.

\section{EXPERIMENTS AND RESULTS} \label{sect:experiments}

\subsection{CNN for camera relocalisation}

In this section we study the relocalisation performance depending on the input type where we use a fixed architecture while the nature of the input varies. For ease, we use a fast implementation; more specifically, the VGG-F architecture is tested with pre-trained weights optimized for object detection in the ImageNet dataset \cite{deng:dong:2009} as well as with randomly initialised weights, and we evaluated them in the TUM's long household and office sequence \cite{sturm:2012}, as no significant difference between architectures has been observed in this scene and therefore the difference in performance is attributable to the difference in the input.

\begin{figure}[h]
\centering
	\begin{subfigure}{.15\textwidth}
  		\centering
  		\includegraphics[width=1\textwidth]{{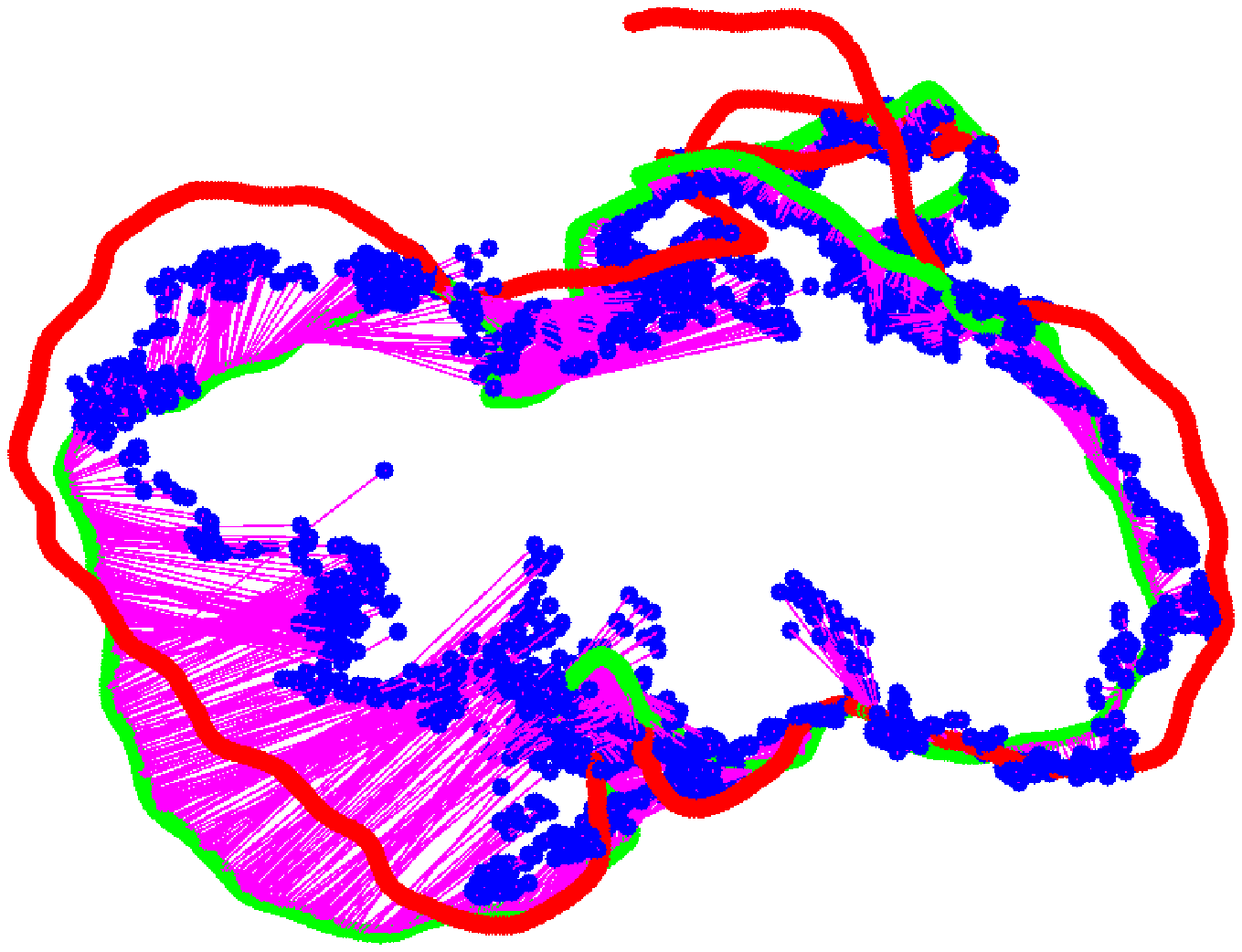}}
  		\caption{}
  		\label{fig:tumrel1}
	\end{subfigure}
	\begin{subfigure}{.15\textwidth}
  		\centering
  		\includegraphics[width=1\textwidth]{{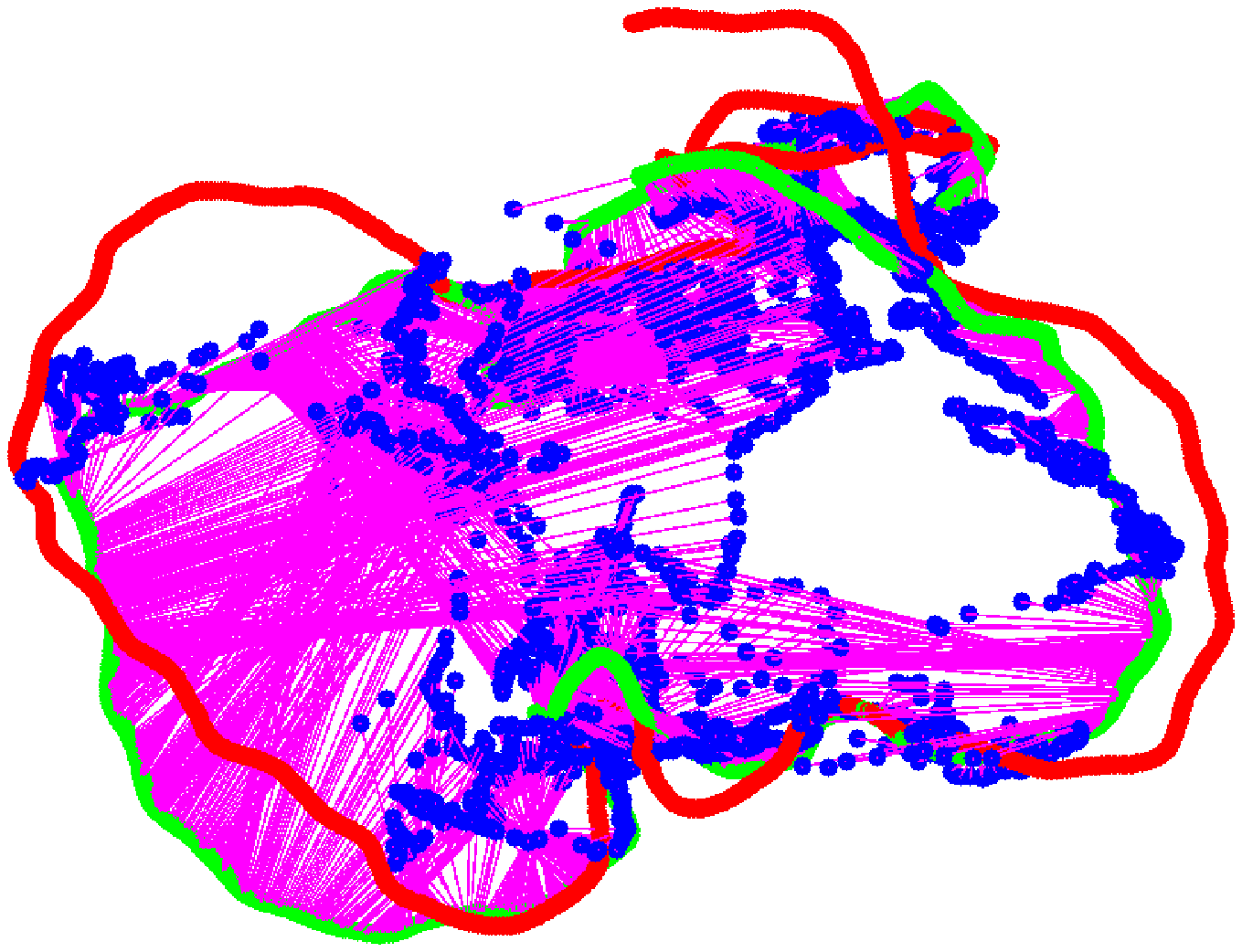}}
  		\caption{}
  		\label{fig:tumrel2}
	\end{subfigure}
	\begin{subfigure}{.15\textwidth}
  		\centering
  		\includegraphics[width=1\textwidth]{{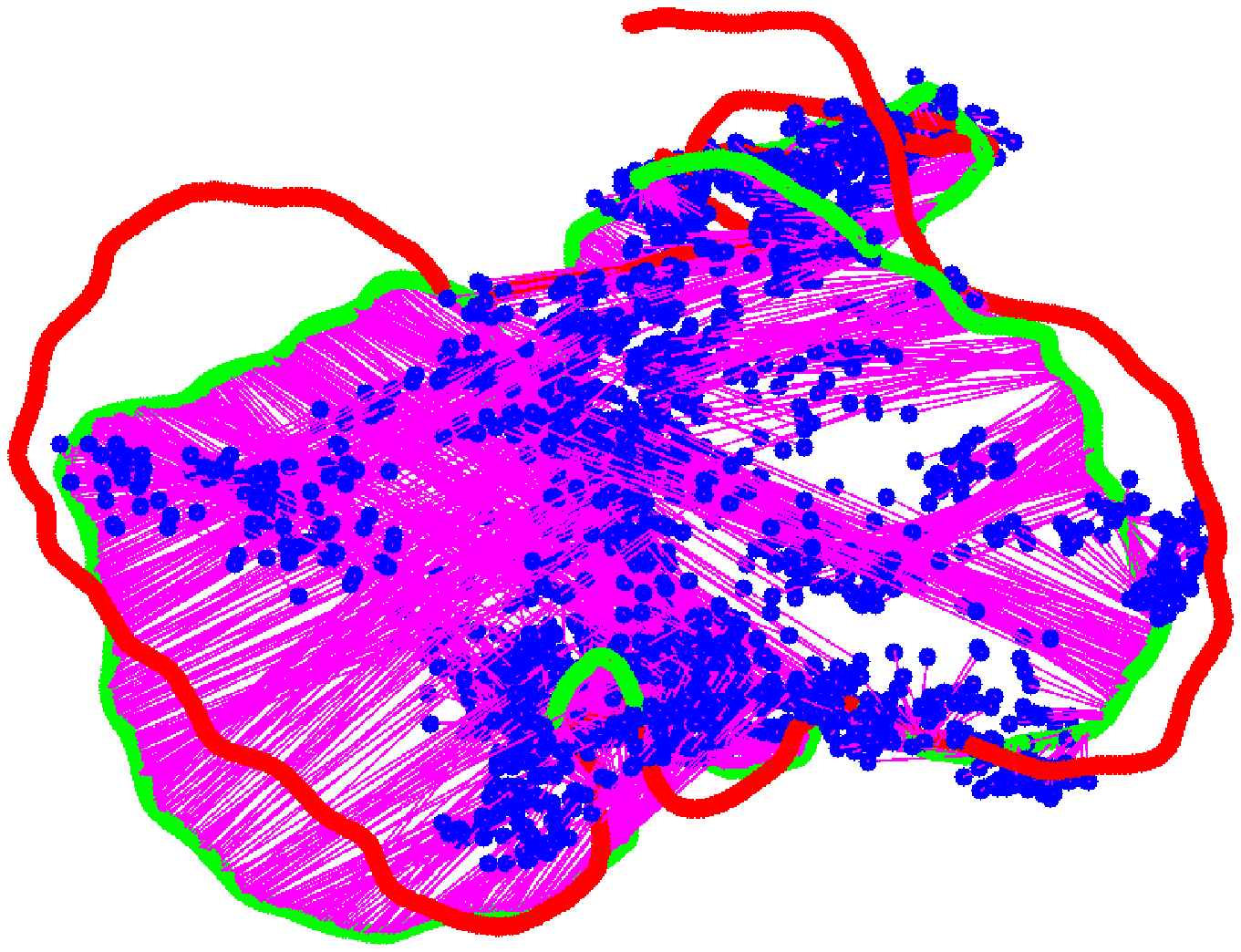}}
  		\caption{}
  		\label{fig:tumrel3}
	\end{subfigure}
	\caption{Relocalisation performance in the TUM sequence \cite{sturm:2012} and a) an RGB input with a pre-trained CNN in the ImageNet dataset \cite{deng:dong:2009}, b) an RGB input and a randomly initialized CNN, and c) a point cloud of 3D points input. Red line indicates the training trajectory while the green line is the testing one. Blue points are the neural network's output.}
	\label{fig:tumrel}
\end{figure}

First, the response to an RGB input can be seen in Figure \ref{fig:tumrel}, where the red line indicates the training sequence and the green line the test one -- results show a relocalisation mean error of 0.572 meters using pre-trained weights (Figure \ref{fig:tumrel1}), and 0.867 meters when random weights were used (Figure \ref{fig:tumrel2}), where the difference in performance is attributable to the filters initialisation.

Then, we evaluate other sensor inputs with a different nature than RGB data, e.g. Figure \ref{fig:tumrel3} shows the relocalisation performance for a dense point cloud of 3D points. The CNN weights were randomly initialized, as pre-trained CNNs are more common for RGB than depth or spatial data.

Table \ref{tab:cnntum} shows the relocalisation mean error for all different inputs after 1000 epochs. Again, the pre-trained CNN on RGB information seems to outperform the others; however, with exception of the pre-trained cases, there is not a significant difference among the distinct input data, and this is attributable to the lack of training samples. We also can observe that, when combined information layers are used, the performance decreases, what might be due to the difference in the input nature (color, depth, and spatial position). One way to overcome it can be the use of parallel networks for each input and then average the output, as in \cite{agrawal:2015}.

\begin{table}[htp]
\centering
	\caption{CNN-F relocalisation mean error [in meters] for different inputs after 1000 epochs in the long office sequence from the TUM dataset \cite{sturm:2012}.}
	\label{tab:cnntum}
	\centering
	\resizebox{0.45\textwidth}{!}{
		\begin{tabular}{|c|c|c|}
		\hline
			&\multicolumn{2}{|c|}{Relocalisation mean error}\\ 
		\cline{2-3}
			Input&Position [m]& Angle [$^\circ$]\\
		\hline
			Depth & 1.768 $\pm$ 0.568 & 44.93 $\pm$ 32.78\\
			Gray & 0.832 $\pm$ 0.675 &  31.91 $\pm$ 42.13 \\
            Point Cloud & 0.986 $\pm$ 0.834 & 39.14 $\pm$ 46.85\\
            RGB & 0.778 $\pm$ 0.675 & 27.39 $\pm$ 41.43\\
            \textbf{Pre-trained RGB} & \textbf{0.465 $\pm$ 0.447} & \textbf{22.16 $\pm$ 40.91}\\
            RGB+Depth & 0.863 $\pm$ 0.730 & 28.68 $\pm$ 41.67\\
            RGB+Point Cloud & 2.330 $\pm$ 0.494 & 79.63$\pm$ 24.39 \\
		\hline
		\end{tabular}}
\end{table}

\begin{table*}[htp]
\centering
	\caption{Relocalisation mean error [in meters] for different architectures in several datasets given an RGB input.}
	\label{tab:posalldatasets}
	\centering
	\resizebox{0.95\textwidth}{!}{
		\begin{tabular}{|c|c|c|c|c|c|}
		\hline
			&PoseNet&TUM&\multicolumn{3}{|c|}{7Scenes}\\ 
		\cline{2-6}
			&St Marys Church&Long Office&Pumpkin&Red Kitchen&Office\\
		\hline
			VGG-F & 8.061 $\pm$ 6.193 & 0.541 $\pm$ 0.342 & 0.575 $\pm$ 0.278 & 0.616 $\pm$ 0.430 & 0.478 $\pm$ 0.258 \\
			VGG-M & 8.282 $\pm$ 6.489 & 0.554 $\pm$ 0.384 & 0.606 $\pm$ 0.299 & 0.590 $\pm$ 0.453 & 0.489 $\pm$ 0.276 \\
            VGG-S & 8.784 $\pm$ 6.694 & 0.544 $\pm$ 0.359 & 0.608 $\pm$ 0.315 & 0.628 $\pm$ 0.454 & 0.521 $\pm$ 0.302 \\
            VGG-16 & 4.671 $\pm$ 4.419 & 0.468 $\pm$ 0.367 & 0.448 $\pm$ 0.272 & 0.483 $\pm$ 0.352 & 0.345 $\pm$ 0.197 \\
            VGG-19 & 4.760 $\pm$ 4.620 & 0.470 $\pm$ 0.362 & 0.446 $\pm$ 0.264 & 0.471 $\pm$ 0.372 & 0.350 $\pm$ 0.217 \\
            \textit{PoseNet} \cite{kendall:grimes:2015} & 2.65 & NA & 0.47 & 0.59 & 0.48 \\
            \textit{SCoRe Forest} \cite{shotton:glocker:2013} & NA & NA & 0.04 & 0.04 & 0.04 \\
            \textit{ORB-SLAM2} \cite{murartal:2017} & NA & 0.01 & NA & NA & NA \\
		\hline
		\end{tabular}}
\end{table*}

\begin{table*}[htp]
\centering
	\caption{Relocalisation mean error [in degrees] for different architectures in several datasets and an RGB input.}
	\label{tab:angalldatasets}
	\centering
	\resizebox{0.95\textwidth}{!}{
		\begin{tabular}{|c|c|c|c|c|c|}
		\hline
			&PoseNet&TUM&\multicolumn{3}{|c|}{7Scenes}\\ 
		\cline{2-6}
			&St Marys Church&Long Office&Pumpkin&Red Kitchen&Office\\
		\hline
			VGG-F & 13.25 $\pm$ 15.14 & 25.63 $\pm$ 44.68 & 9.67 $\pm$ 6.89 & 10.67 $\pm$ 9.24 & 10.66 $\pm$ 7.90 \\
			VGG-M & 12.97 $\pm$ 16.57 & 24.72 $\pm$ 39.82 & 9.04 $\pm$ 6.79 & 10.82 $\pm$ 8.68 & 11.07 $\pm$ 7.71 \\
            VGG-S & 14.18 $\pm$ 19.30 & 25.99 $\pm$ 41.96 & 9.72 $\pm$ 8.39 & 11.14 $\pm$ 11.82 & 11.76 $\pm$ 8.41 \\
            VGG-16 & 9.84 $\pm$ 16.59 & 28.96 $\pm$ 43.46 & 9.59 $\pm$ 7.08 & 8.45 $\pm$ 7.75 & 8.35 $\pm$ 7.10 \\
            VGG-19 & 10.41 $\pm$ 17.40 & 25.68 $\pm$ 42.19 & 8.88 $\pm$ 7.41 & 8.10 $\pm$ 7.57 & 9.10 $\pm$ 8.27 \\
            \textit{PoseNet} \cite{kendall:grimes:2015} & 4.24 & NA & 4.21 & 4.32 & 3.84 \\
            \textit{SCoRe Forest} \cite{shotton:glocker:2013} & NA & NA & 0.68 & 0.76 & 0.78 \\
            \textit{ORB-SLAM2} \cite{murartal:2017} & NA & NA & NA & NA & NA \\
		\hline
		\end{tabular}}
\end{table*}

\subsection{Multiple trajectories learning}

\begin{figure}[h]
\centering
	\begin{subfigure}{.1125\textwidth}
		\centering
		\includegraphics[bb = 0 0 640 480, width=1\textwidth]{{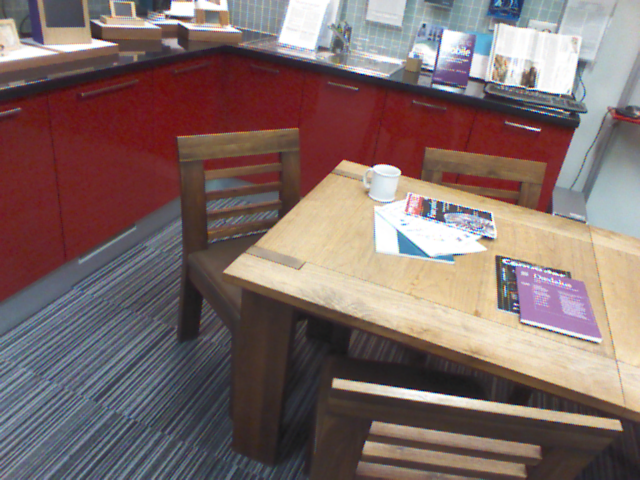}}
		\caption{}
		\label{fig:7scenes1}
	\end{subfigure}
	\begin{subfigure}{.1125\textwidth}
  		\centering
  		\includegraphics[bb = 0 0 640 480, width=1\textwidth]{{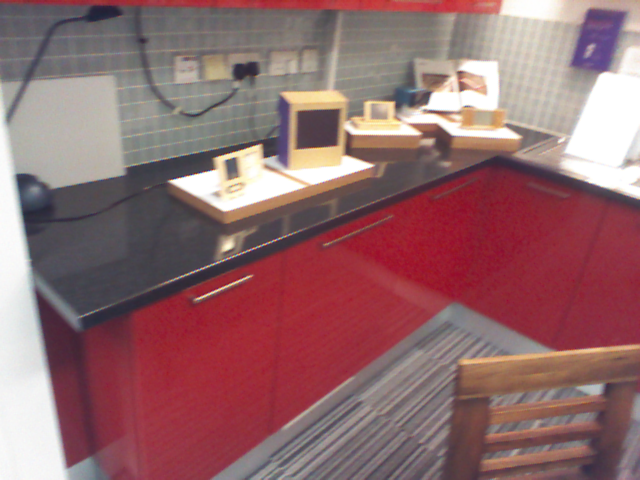}}
  		\caption{}
  		\label{fig:7scenes2}
	\end{subfigure}
	\begin{subfigure}{.1125\textwidth}
  		\centering
  		\includegraphics[bb = 0 0 640 480, width=1\textwidth]{{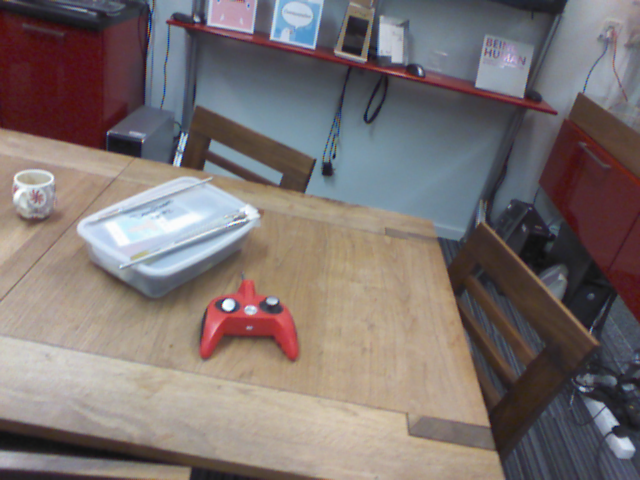}}
  		\caption{}
  		\label{fig:7scenes3}
	\end{subfigure}
	\begin{subfigure}{.1125\textwidth}
  		\centering
  		\includegraphics[bb = 0 0 640 480, width=1\textwidth]{{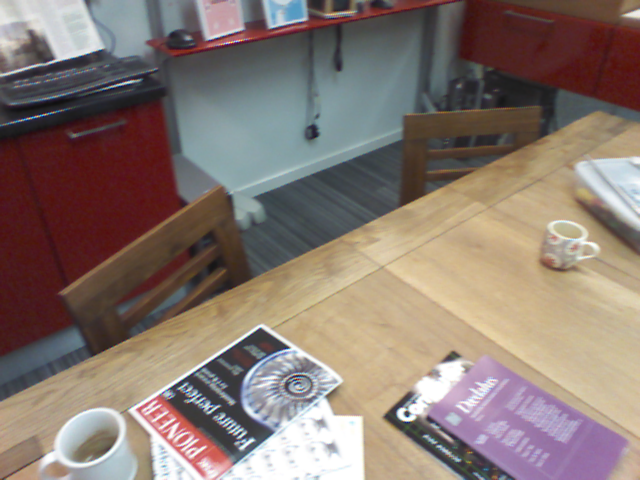}}
  		\caption{}
  		\label{fig:7scenes4}
	\end{subfigure}
	\caption{Typical views from the Red Kitchen sequence in the 7-Scenes dataset \cite{glocker:izadi:2013}.}
	\label{fig:7scenes}
\end{figure}

To test the relocalisation performance with respect to the number of training sequences, and hence the neural network capability for map representation and compression, we use the Microsoft's 7-Scenes dataset (\cite{glocker:izadi:2013}, \cite{shotton:glocker:2013}), as shown in Figure \ref{fig:7scenes}; this dataset consists of several trajectories taken by different persons moving around the same environment. Training, validation and testing sequences are indicated in the dataset itself.

Similar to the previous section, we evaluate the VGG-F architecture with all different input data and compared their response against PoseNet \cite{kendall:grimes:2015}; results are shown in Table \ref{tab:cnn7scenes}. Although RGB error is the lowest again, in this case, where more training data per set is present, similar performances (within the variance error) are found among different data types. Therefore, without lost of generality, for the rest of the section we will use only RGB-only data.

Additionally, from Table \ref{tab:cnn7scenes} we observe that, in the case of RGB data, the VGG-F (8 layers) behaves as good as PoseNet, a 23 layers and more complex neural network, with a relocalisation mean error of 0.559 meters in the former and 0.59 meters in the latter, and therefore a compression in the number of layers is achieved. It remains an open problem the task of designing customized CNN map representation by systematically modifying the neural network architecture itself.

\begin{table}[htp]
\centering
	\caption{Relocalisation mean error [in meters] using VGG-F for different inputs after 1000 epochs in the Red Kitchen sequence from the 7-Scenes dataset\cite{glocker:izadi:2013}. PoseNet mean error is indicated in italics, as reported in \cite{kendall:grimes:2015}.}
	\label{tab:cnn7scenes}
	\centering
	\resizebox{0.45\textwidth}{!}{
		\begin{tabular}{|c|c|c|}
		\hline
			&\multicolumn{2}{|c|}{Relocalisation mean error}\\ 
		\cline{2-3}
			Input&Position [m]& Angle [$^\circ$]\\
		\hline
			Depth & 1.261 $\pm$ 0.382 & 20.24 $\pm$ 12.21\\
			Gray & 0.747 $\pm$ 0.493 & 12.37 $\pm$ 11.12\\
            Point Cloud & 0.740 $\pm$ 0.666 & 14.11 $\pm$ 13.94\\
            \textbf{RGB} & \textbf{0.559 $\pm$ 0.417} & \textbf{8.57 $\pm$ 7.86}\\
            \textit{PoseNet (RGB)} & \textit{0.59} & \textit{4.32} \\
            RGB+Depth & 0.704 $\pm$ 0.538 & 11.77 $\pm$ 11.22\\
            RGB+Point Cloud & 0.640 $\pm$ 0.661 & 12.12 $\pm$ 13.92\\
		\hline
		\end{tabular}}
\end{table}

The different architectures evaluated in this work are presented in Table \ref{tab:allarchs}. As a reference, it is presented {SCoRe Forest} \cite{shotton:glocker:2013}, a regression forest trained for pixel to 3D point correspondence prediction -- the authors used 5 trees with 500 images per tree and 5000 example pixels per image. It is also presented ORB-SLAM2 \cite{murartal:2017}, where a map with 16k features was generated in the TUM's long office scene \cite{sturm:2012}.

\begin{table}[htp]
\centering
    \caption{Relocalisation mean error [in degrees] for different architectures in several datasets. PoseNet has fewer parameters in a more complex architecture (parameter estimation based on the GoogLeNet architecture \cite{emer:2017}).}
    \label{tab:allarchs}
    \centering
    \resizebox{0.45\textwidth}{!}{
        \begin{tabular}{|c|c|c|}
        \hline
            Model & Parameters & Layers\\ 
        \hline
            VGG-F & 61M & 8 \\
            VGG-M & 100M & 8 \\
            VGG-S & 100M & 8 \\
            VGG-16 & 138M & 16 \\
            VGG-19 & 140M & 19 \\
            \textit{PoseNet} \cite{kendall:grimes:2015} & 7M & 23 \\
            \textit{SCoRe Forest} \cite{shotton:glocker:2013} & 12.5M & NA \\
            \textit{ORB-SLAM2} \cite{murartal:2017} & 16k & NA \\
        \hline
        \end{tabular}}
\end{table}

Table \ref{tab:posalldatasets} and Table \ref{tab:angalldatasets} present a summary of the relocalisation response in position and orientation, respectively, for the best combinations in each architecture using different datasets and an RGB input; again, from an expected pose $p = [x, q]$ and the CNN response $\hat{p} = [\hat{x}, \hat{q}]$, the error in position is given by $e_p = \Vert x - \hat{x}\Vert$, while the error in orientation is given by $ e_a = cos^{-1}(<q, \hat{q}>)$. These results confirm that, when enough training data is available, there is not significant difference in performance among short and long CNN architectures and, therefore, they may be used alike. In addition, it is observed that, with a relatively small regression map (12M in the SCoRe Forest), it is possible to obtain a high accuracy -- as a CNN can approximate any function, it remains an open question how to find the best CNN architecture that approximate those results.

Furthermore, in traditional mapping techniques, the map usually increases when new views are added; instead, when using a CNN map representation, map information increases while maintaining a neural network of constant size by re-training it when new information is added.

\begin{figure}[h]
	\centering
	\includegraphics[width=0.45\textwidth]{{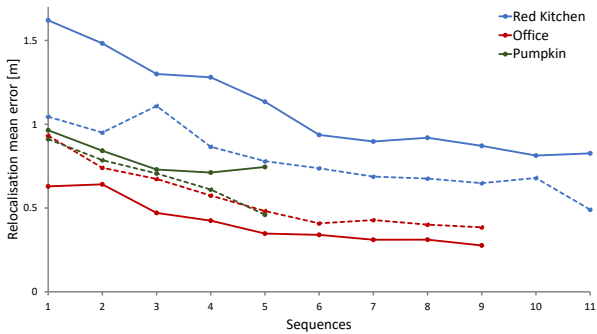}}
	\caption{Relocalisation mean error in several scenes from the 7-Scenes dataset \cite{glocker:izadi:2013} with respect to the number of training trajectories (one sequence is fixed for testing and the remaining are used for training; dotted lines indicates a different test sequence). The VGG-F is utilized for evaluation. While the number of training trajectories increases, the error decreases but the neural network size remains the same (the training only affects the weights).}
	\label{fig:7scenescumul}
\end{figure}

\begin{figure}[h]
	\centering
	\includegraphics[width=0.45\textwidth]{{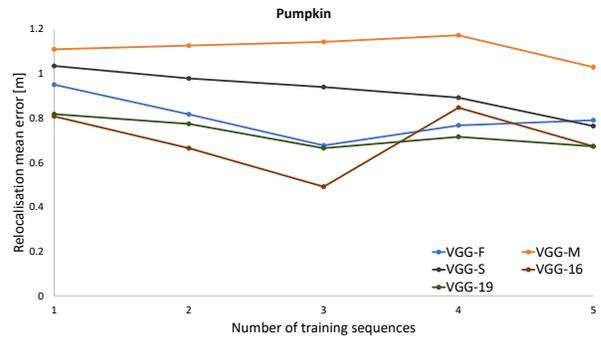}}
	\caption{Relocalisation mean error in the Pumpkin dataset \cite{glocker:izadi:2013} using several CNN architectures with respect to the number of sequences in the training process, except for one sequence selected for testing. No significant difference in the performance between the short (in particular VGG-F) and long architectures is observed and therefore short architectures can be used without sacrificing performance.}
	\label{fig:7scenescumlpumkin}
\end{figure}

\begin{figure}[h]
	\centering
	\includegraphics[width=0.45\textwidth]{{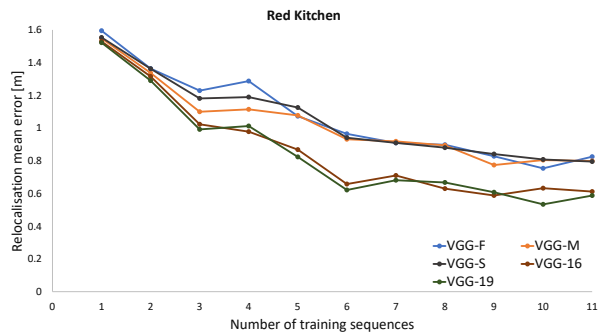}}
	\caption{Relocalisation mean error in the Red Kitchen dataset \cite{glocker:izadi:2013} as in Figure \ref{fig:7scenescumlpumkin}.}
	\label{fig:7scenescumlredkitchen}
\end{figure}

\begin{figure}[h]
	\centering
	\includegraphics[width=0.45\textwidth]{{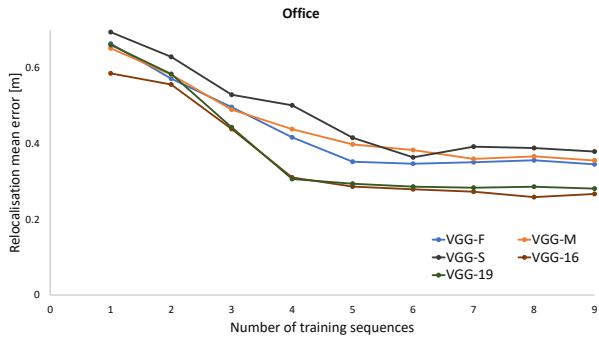}}
	\caption{Relocalisation mean error in the Office dataset \cite{glocker:izadi:2013} as in Figure \ref{fig:7scenescumlpumkin} and Figure \ref{fig:7scenescumlredkitchen}.}
	\label{fig:7scenescumloffice}
\end{figure}

We use the Red Kitchen, Office and Pumpkin sequences in the Microsoft's 7-Scenes dataset to test the CNN saturation point as follows. One of the trajectories is left out for testing, and the CNN is gradually trained by adding one remaining sequence at a time. Figure \ref{fig:7scenescumul} shows that while increasing the number of trajectories, precision also increases but, by construction, the size of the CNN remains the same, as expected. Similarly, Figure \ref{fig:7scenescumlpumkin}, Figure \ref{fig:7scenescumlredkitchen}, and Figure \ref{fig:7scenescumloffice} show the performance in this incremental setup for short and long architectures; no significant difference in relocalisation performance between CNN architectures is observed, and therefore the use of compact CNN map representation in such scenarios is possible. 

Nevertheless, an asymptotic behaviour has not been reached after using all the training sequences, indicating that the neural network is not saturated and suggesting that more trajectories can still be added, improving the performance. While compact, then, this map representation is also constant-size when new information is added. As stated before, compression comes in the sense of finding the smallest neural network architecture with that still performs well in the relocalisation problem. On the other hand, compression also comes in the sense that the map information increases when new information is received without increasing the map representation size given by a constant-size CNN architecture.

\begin{figure}[ht]
\centering
	\begin{subfigure}{.1125\textwidth}
		\centering
		\includegraphics[width=1\textwidth]{{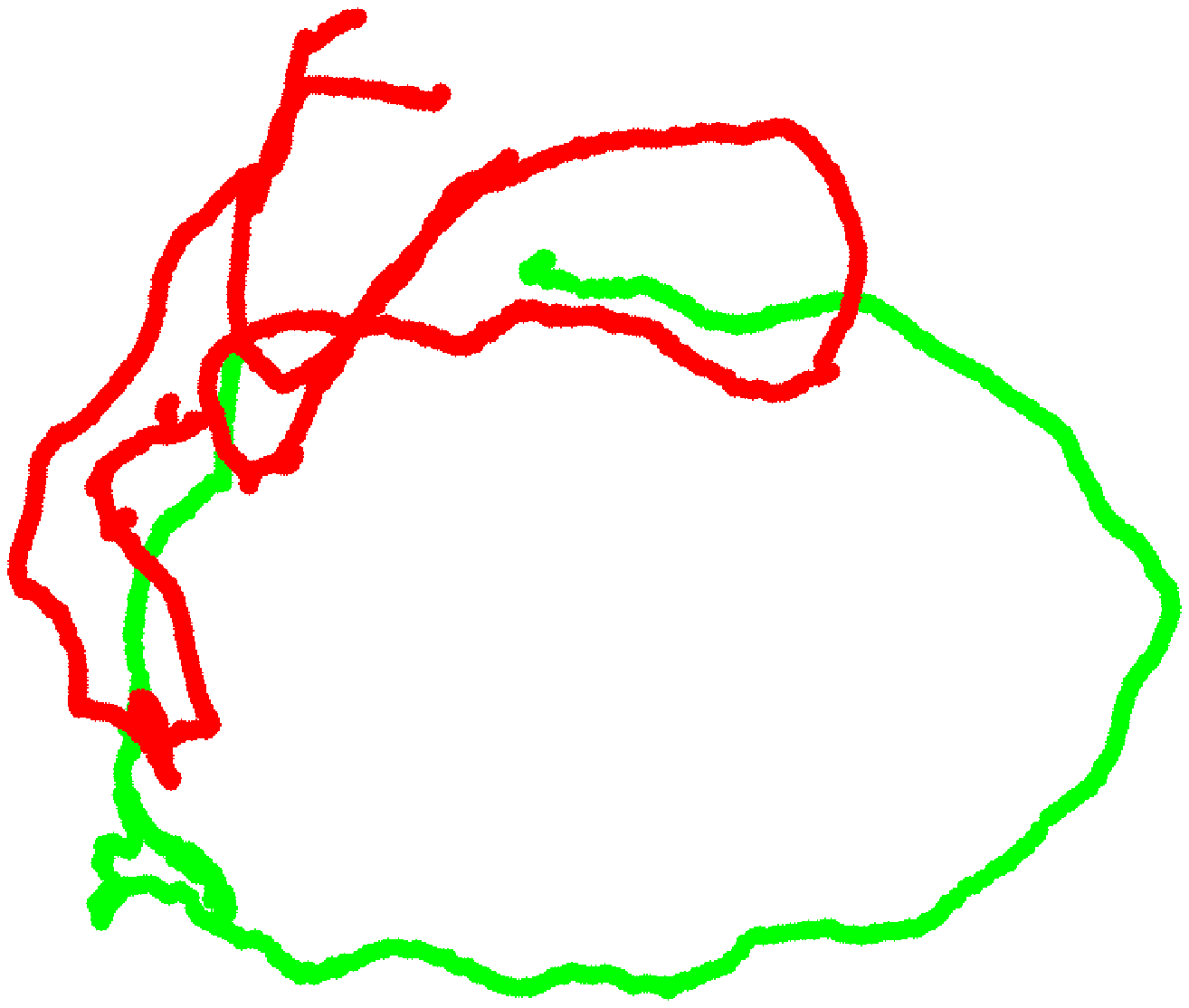}}
		\label{fig:redkitchen2a}
	\end{subfigure}
	\begin{subfigure}{.1125\textwidth}
  		\centering
  		\includegraphics[width=1\textwidth]{{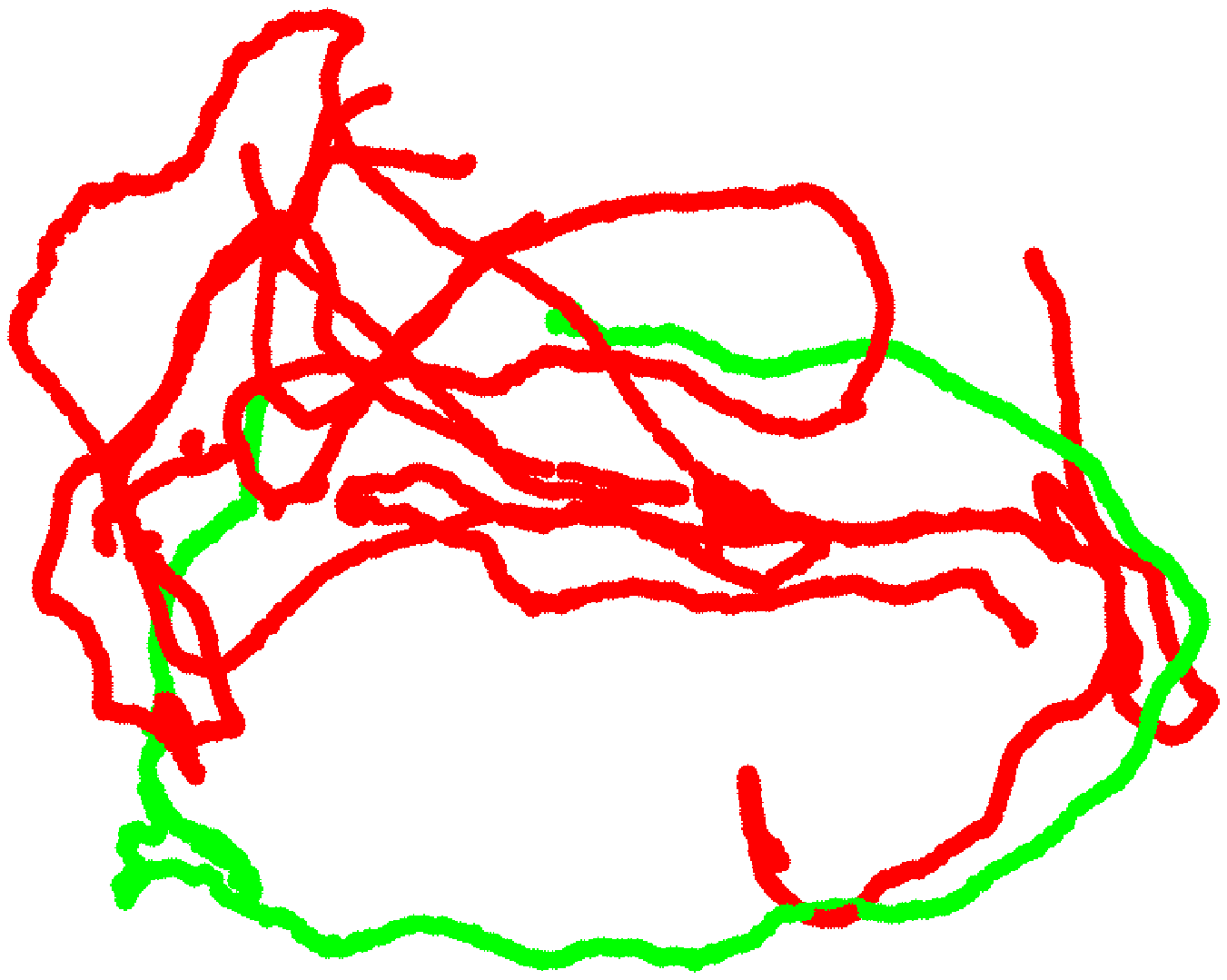}}
  		\label{fig:redkitchen5a}
	\end{subfigure}
	\begin{subfigure}{.1125\textwidth}
  		\centering
  		\includegraphics[width=1\textwidth]{{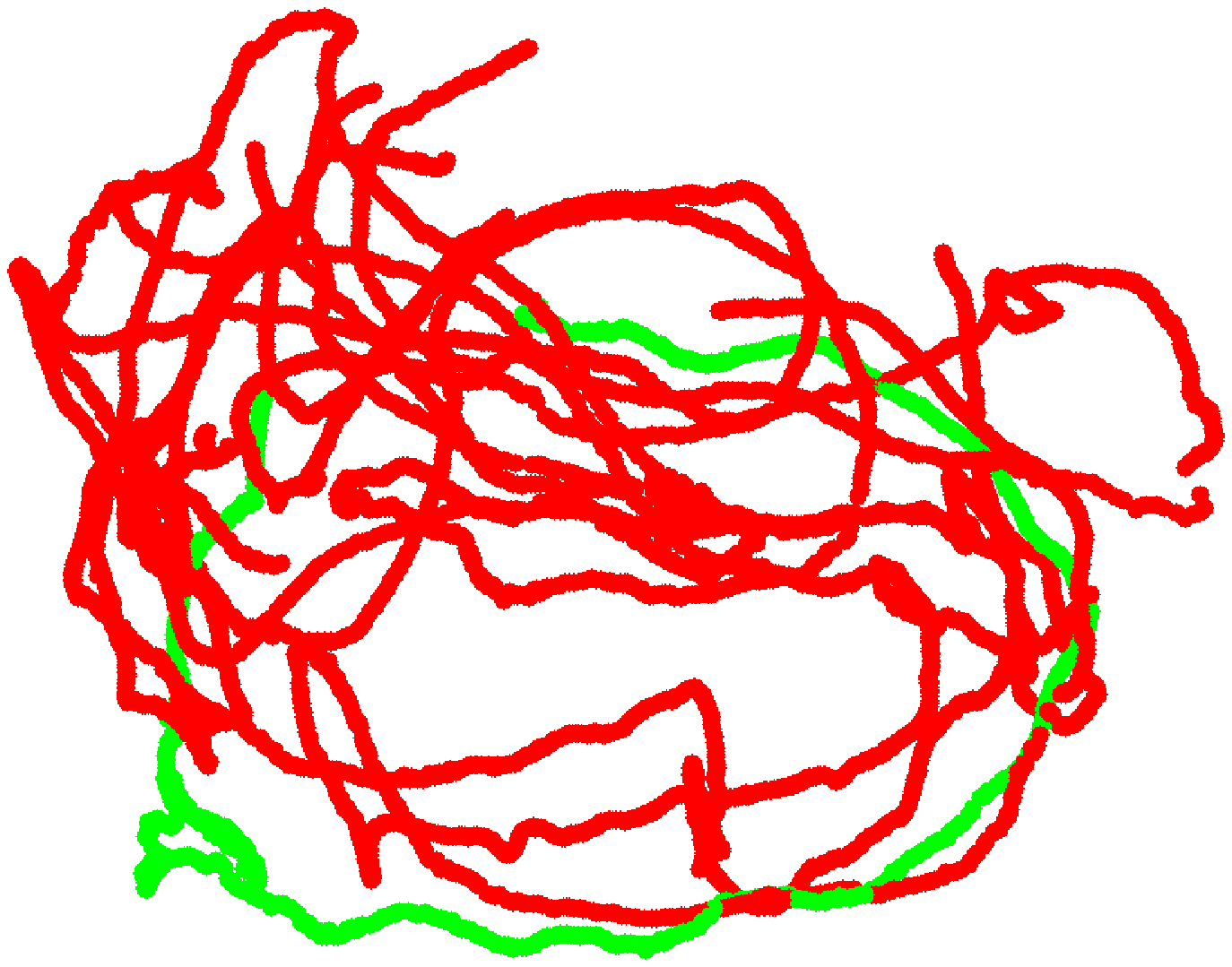}}
  		\label{fig:redkitchen8a}
	\end{subfigure} 
	\begin{subfigure}{.1125\textwidth}
  		\centering
  		\includegraphics[width=1\textwidth]{{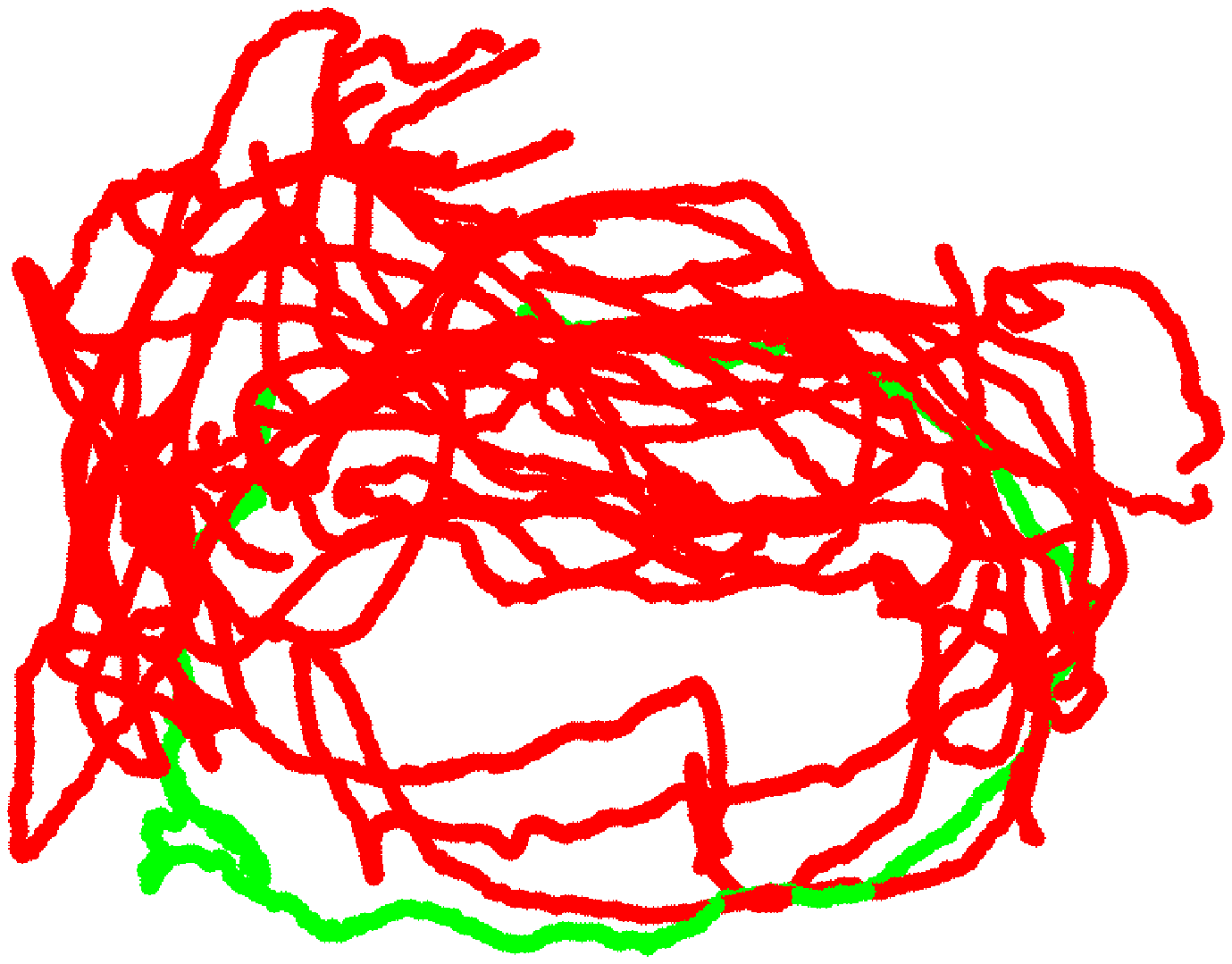}}
  		\label{fig:redkitchen11a}
	\end{subfigure} \\
	\begin{subfigure}{.1125\textwidth}
		\centering
		\includegraphics[width=1\textwidth]{{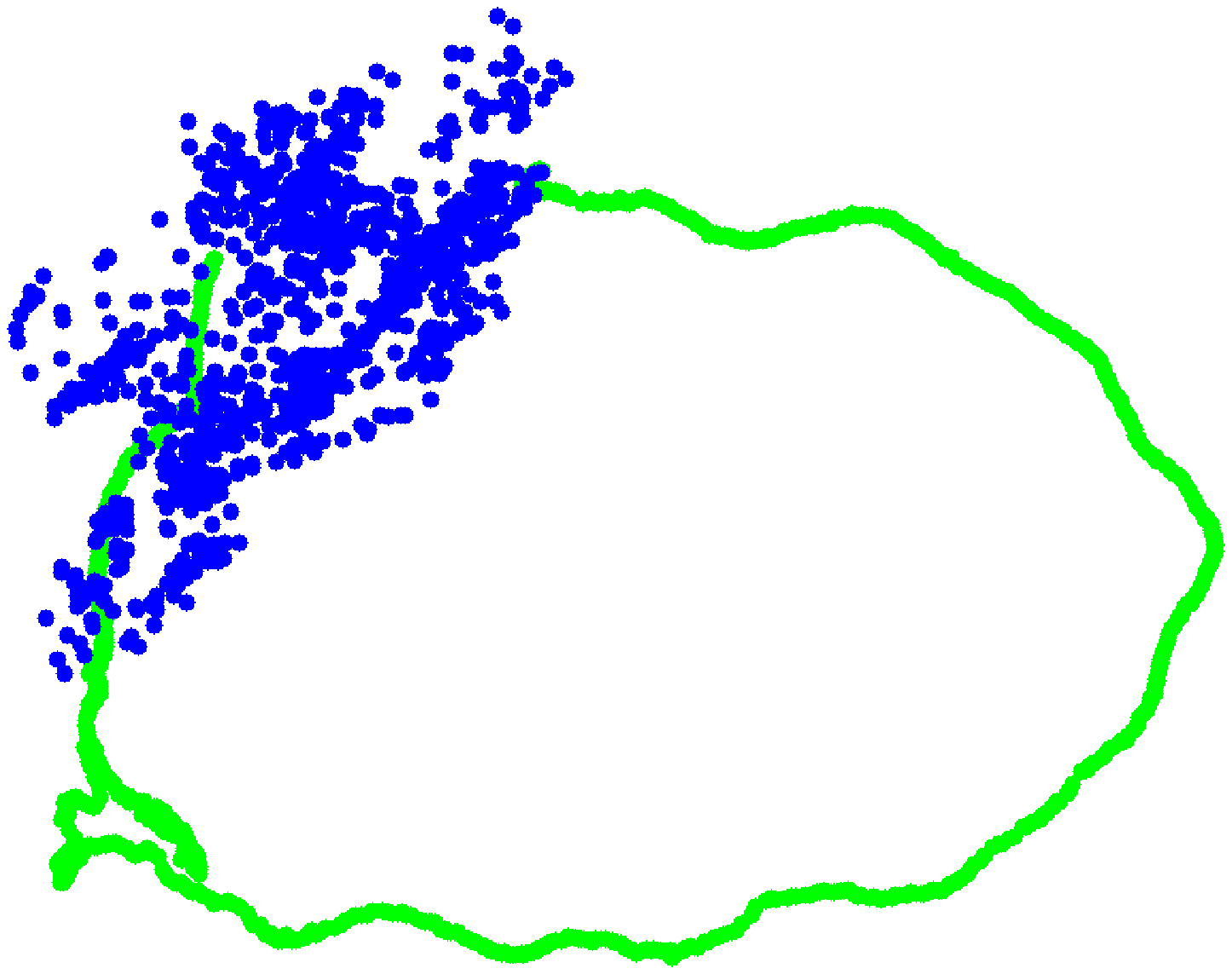}}
		\caption{}
		\label{fig:redkitchen2b}
	\end{subfigure}
	\begin{subfigure}{.1125\textwidth}
  		\centering
  		\includegraphics[width=1\textwidth]{{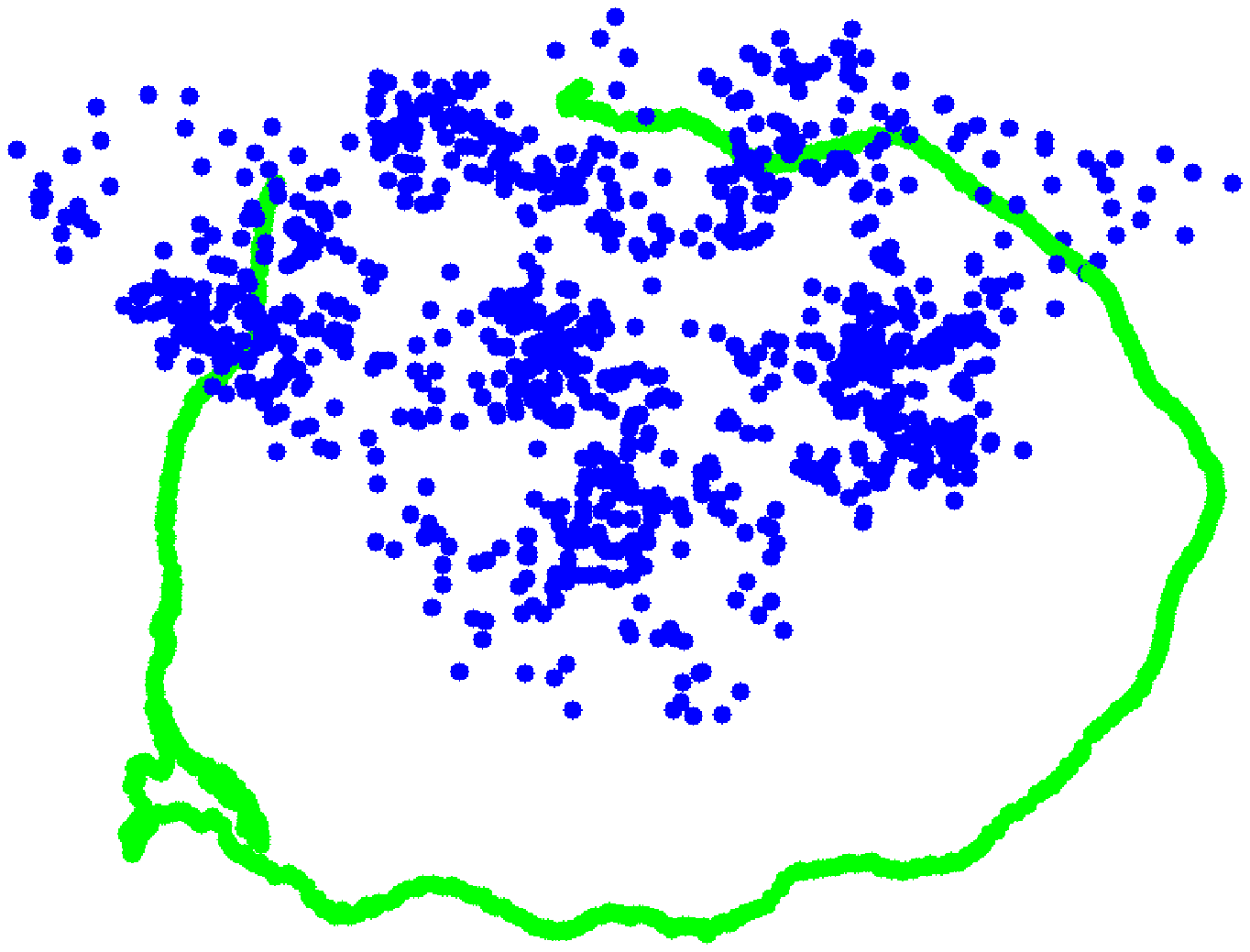}}
  		\caption{}
  		\label{fig:redkitchen5b}
	\end{subfigure}
	\begin{subfigure}{.1125\textwidth}
  		\centering
  		\includegraphics[width=1\textwidth]{{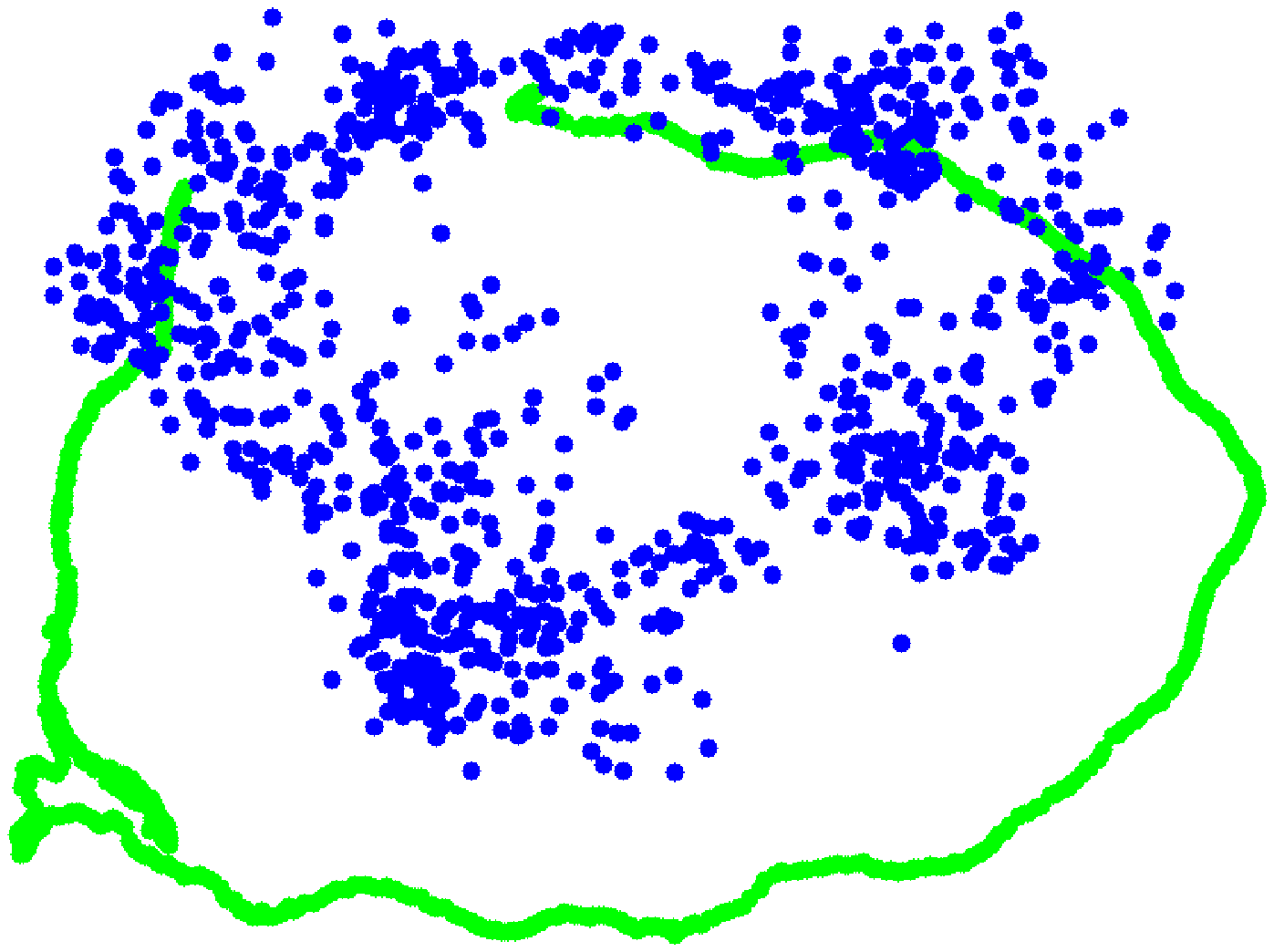}}
  		\caption{}
  		\label{fig:redkitchen8b}
	\end{subfigure}
	\begin{subfigure}{.1125\textwidth}
  		\centering
  		\includegraphics[width=1\textwidth]{{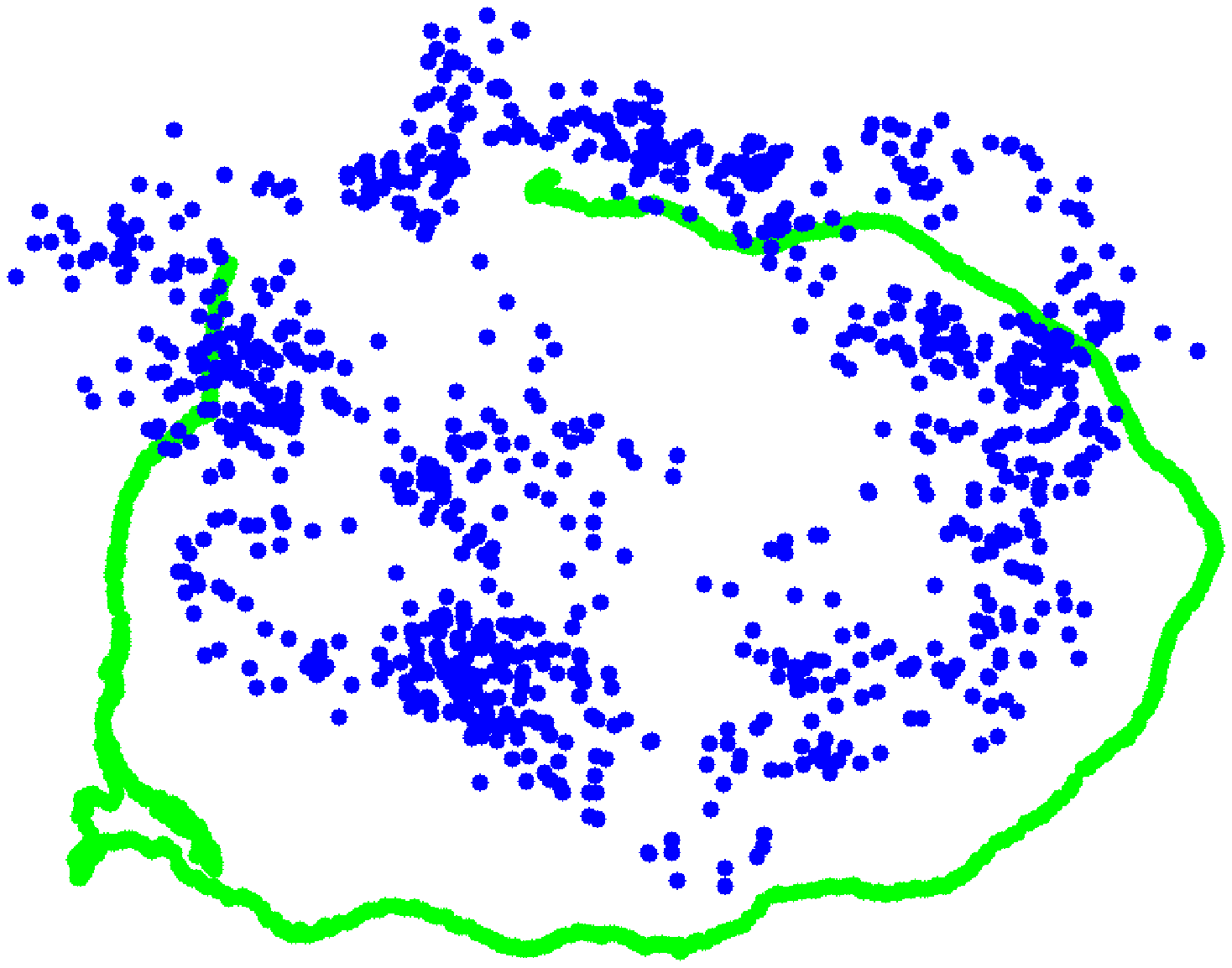}}
  		\caption{}
  		\label{fig:redkitchen11b}
	\end{subfigure} 
	\caption{Relocalisation performance for the Red Kitchen sequence in the Microsoft-s 7-Scenes dataset \cite{glocker:izadi:2013} after training with a) two, b) five, c) eight, and d) eleven sequences using the VGG-F. Red lines indicate the training sequences and the green line is the test one; blue points are the output of the system.}
	\label{fig:allredkitchen}
\end{figure}

Figure \ref{fig:allredkitchen} shows some outputs for the Red Kitchen sequence where the relocalisation improves as more trajectories are used using the VGG-F fast architecture. There, it is also observed how the relocated cameras (blue points) are closer to the test sequence (green line) when more training sequences are present (red line).

\section{Conclusions}

We presented a first approach toward CNN map representation and compression for camera relocalisation. The response to different inputs and to different trajectories was studied. We first shown that for these kind of models, when few training data (some thousands samples) and training from scratch, the RGB images present the best performance compared with other types of data, as depth or 3D point clouds. Then, we presented the idea of map compression as the task of finding optimal CNN architectures where different architectures were evaluated: VGG-F, VGG-M, and VGG-S, all with 8 layers and different orders of complexity, VGG-16 (16 layers), and VGG-19 (19 layers). We observed that, when the amount of training data is reduced, the architecture plays a crucial role in the relocalisation performance; however, with a representative training set, the short and long architectures perform similarly. None o those architectures outperforms state-of-the-art techniques (e.g. ORB-SLAM); however, regression techniques like SCoRe Forests show promising results for map representations as regression models.

On the other hand, we perform a study on the relocalisation performance with respect to the number of training sequences. We observed that the performance increases with the number of training sequences, as expected. However, in the context of map representation through neural networks, it means that the map accuracy can be improved without increasing the map size but only by adjusting its weights. This is important when a fixed amount of memory is available to store the map in the mapping and navigation process.

For future work we note that more complex relocalisation such as semantic or topological relocalisation were not explored here. One potential direction encouraged by these results is to train simpler networks for object recognition with labeled datasets, and use a second network that accepts semantic information as input for relocalisation. This kind of multi-network systems, where two complex systems interact to perform a single task are of interest to expand on the current work.

\addtolength{\textheight}{-15cm} 









\bibliographystyle{ieeetr}
\bibliography{all}

\end{document}